\documentclass{article}

\usepackage{microtype}
\usepackage{graphicx}
\usepackage{subfigure}
\usepackage{subcaption}
\usepackage{booktabs} 
\usepackage{xcolor}

\usepackage{hyperref}

\usepackage[accepted]{icml2024}
\usepackage{amsmath}
\usepackage{amssymb}
\usepackage{mathtools}
\usepackage{amsthm}

\usepackage[capitalize,noabbrev]{cleveref}

\usepackage[inkscapelatex=false]{svg}
\theoremstyle{plain}

\theoremstyle{definition}

\theoremstyle{remark}

\usepackage[disable,textsize=tiny]{todonotes}

\icmltitlerunning{Bilinear Conv Decomposition for Causal RL Interpretability}

\begin{document}

\twocolumn[
\icmltitle{Bilinear Convolution Decomposition for Causal RL Interpretability}

\icmlsetsymbol{equal}{*}

\begin{icmlauthorlist}
\icmlauthor{Narmeen Oozeer}{equal,yyy}
\icmlauthor{Sinem Erisken}{equal,yyy}
\icmlauthor{Alice Rigg}{equal,yyy}

\end{icmlauthorlist}

\icmlaffiliation{yyy}{Independent}

\icmlcorrespondingauthor{Alice Rigg}{rigg.alice0@gmail.com}

\vskip 0.3in
]

\begin{abstract}
Efforts to interpret reinforcement learning (RL) models often rely on high-level techniques such as attribution or probing, which provide only correlational insights and coarse causal control.
This work proposes replacing nonlinearities in convolutional neural networks (ConvNets) with bilinear variants, to produce a class of models for which these limitations can be addressed. We show bilinear model variants perform comparably in model-free reinforcement learning settings, and give a side by side comparison on ProcGen environments. Bilinear layers' analytic structure enables weight-based decomposition. Previous work has shown bilinearity enables quantifying functional importance through eigendecomposition, to identify interpretable low rank structure \cite{pearce2024bilinear}. 
We show how to adapt the decomposition to convolution layers by applying singular value decomposition to vectors of interest, to separate the channel and spatial dimensions. Finally, we propose a methodology for causally validating concept-based probes, and illustrate its utility by studying a maze-solving agent's ability to track a cheese object.
\end{abstract}

\section{Introduction}
Understanding how objectives are encoded in a model such that we could control its outcomes, could help solve the inner alignment problem. Demonstrating this would require causal understanding. Interpreting reinforcement learning (RL) models remains a core challenge, especially when attribution and probing techniques fall short at providing causal, fine-grained insights into learned behaviors. 

The difficulty of refining the causal relevance of probes remains a challenge for useful applications of RL interpretability. Recently, \cite{bilinear_note} proposed that bilinear MLPs are more interpretable. \cite{pearce2024bilinear} shows this by deriving insights through the weights. In this paper, we adapt weight decomposition to concept based interpretability, and introduce an approach that may help address these challenges. Concretely, our contributions are as follows:
\begin{itemize}
\item We motivate bilinear convolution layers as an interpretable model component in ConvNets. We show how bilinear convolution layers can be decomposed into bases of self interacting eigenfilters.
\item We show that bilinear adaptations of RL agents train well in ProcGen environments. We find comparable performance for a simplified IMPALA and a bilinear variant we call Bimpala.
\item We propose a protocol for analyzing mechanisms in Bimpala, to causally validate and mechanistically interpret concept-based probes.
\item We present preliminary interpretability results in studying an RL maze solving agent, demonstrating causal relevance of linear probes in intermediate layers.
\end{itemize}

\section{Bilinear layers}
In this section we'll discuss standard components in deep learning architectures, and the appropriate bilinear counterparts.

\subsection{MLP}
A conventional multi-layer perceptron (MLP) or fully-connected (FC) layer takes a vector $x$ (the hidden representation), passes it through two learned linear transformations (represented by matrices $W_1, W_2$ and bias vectors $b_1, b_2$). An activation function, such as a rectified-linear unit (ReLU) is applied between the two linear transformations. Adopting the notation used in \citep{shazeer_gated}, the generic structure of an MLP is given by
$$FC_\text{Enc}(x, W_1, W_2, b_1, b_2)=\text{Enc}_\sigma(x,W_1,b_1)W_2 + b_2$$
Where the encoder part of the FC is a nonlinear function $\text{Enc}:\mathbb{R}^n \to \mathbb{R}^m$, and has the general form $$\text{Enc}_\sigma(x,W_1,b_1)=\sigma(xW_1+b_1).$$ Here, the encoder has $m$ neurons, and any individual neuron activation is just the corresponding dimension activation for the encoder. Modern models, such as LLMs, feature an encoder variant called a Gated Linear Unit (GLU), which have the following structure, where bilinear encoders use an identity activation:
$$\text{GLU}(x,W,V,b,c)=\sigma(xW+b)\odot (xV+c)$$
$$\text{Bilinear}(x,W,V,b,c)=(xW+b) \odot (xV+c)$$
Thus, as used in a FC, or MLP, the full parametrization, omitting biases for brevity, is given by
\begin{align*} 
\text{FC}_\text{GLU}(x,F,H,W_2)
&=\text{GLU}(x,F,H)W_2 \\
&=(\sigma(xF)\odot(xH))W_2
\end{align*}
\begin{align} 
\label{eq:fcbilinear}
\text{FC}_\text{Bilinear}(x,F,H,W_2)
&=\text{Bilinear}(x,F,H)W_2 \nonumber \\
&= ((xF)\odot(xH))W_2
\end{align}

In this work, we only consider one fully connected layer, therefore omitting the downward projection $W_2$ from the $\text{FC}_\text{Bilinear}$ equation.

\subsection{Convolutions}
A convolution layer is characterized by integers $K$ for kernel width, as well as integer $\text{stride}$ and $\text{padding}$ values. Convolution layers have the form
$$\text{Conv2D}(X,\mathbf{U})=\sigma(X\ast \mathbf{U})$$
where $X$ has shape $[\text{width} \times \text{height} \times C_\text{in}]$, its output $\text{Conv2D}(X,\mathbf{U})$ has shape $[\frac{\text{width}}{\text{stride}} \times \frac{\text{height}}{\text{stride}} \times C_\text{out}]$, and $\sigma$ is an optional activation function applied point-wise. $\mathbf{U}$, the kernel which captures its learned parameters, has shape $K \times K \times C_\text{in} \times C_\text{out}$, and acts locally on $K \times K$ patches. Let $\ell=\lfloor\frac{K}{2}\rfloor$. Given input coordinates $\alpha, \beta$ and an output channel dimension $i$ to write to, a Conv layer's kernel weights act on a local patch around $(\alpha,\beta)$ via the following expression:
\begin{align*}
u_{\alpha, \beta, i}
&= \sum_j \sum_{k_1}^{K} \sum_{k_2}^{K} \mathbf{U}^{(i)}[j,k_1,k_2] \cdot X[j,\alpha+k_1,\beta+k_2]
\end{align*}
where $j$ is an input channel of the 2D convolution, n is the number of input channels and $K=2\ell$. Note that summing over $K$ means summing over $-\ell$ to $\ell$. 
Similar to GLUs for MLPs, a gated or bilinear convolution can replace a traditional convolution layer with activations:
\begin{align}
\label{eq:bconv2d}
    \text{GConv}(x,U,V)&= \sigma(x \ast U) \odot (x \ast V) \nonumber \\
    \text{BConv2D}(x,U,V)&= (x \ast U) \odot (x \ast V)
\end{align}
\begin{figure}[ht]
\begin{center}
\centerline{\includegraphics[width=\columnwidth]{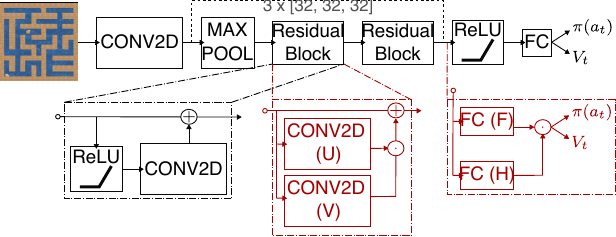}}
\caption{Bimpala: we modified a simplified IMPALA architecture (black) by replacing the operation ReLU(Conv2D) (\autoref{eq:bconv2d}) with BConv2D which consists of gating $2$ Conv2D blocks. We also swap Relu(FC) with FCBilinear (\autoref{eq:fcbilinear}). (red)}
\label{fig:bimpala_architecture}
\end{center}
\end{figure}

\begin{figure*}[t]
\label{eq:FC_bilinear}\includegraphics[width=\textwidth]{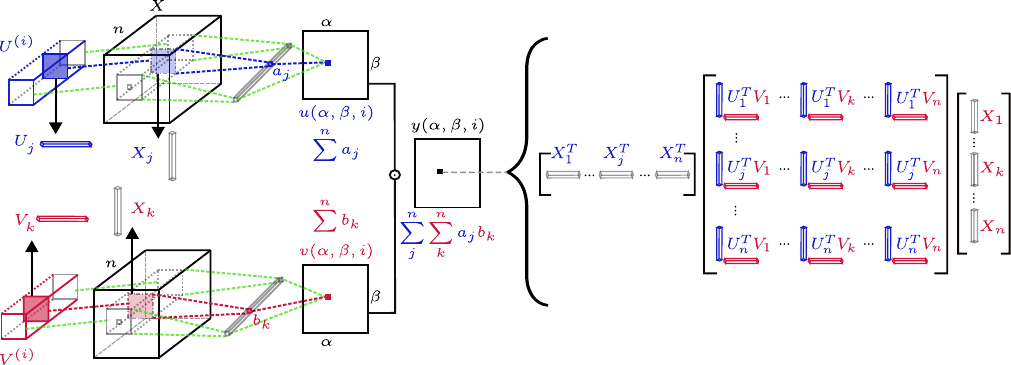}
\caption{Visualization of the quadratic form derivation for gated bilinear convolutions. The diagram illustrates the transformation from spatial convolution operations to a bilinear matrix form in three stages: (1)~The upper path shows the computation of spatial convolutions $U^{(i)}$ with input $X_j$, producing terms $a_j$. (2)~The lower path similarly computes convolutions $V^{(i)}$ with input $X_k$, producing terms $b_k$. (3)~The right side demonstrates how these operations can be reformulated as a product of three block matrices, where the outer product of channel responses $(U^\top V)$ forms a symmetric bilinear matrix. The diagram emphasizes how local spatial convolutions (shown in the cubes) are transformed into a bilinear form $B$.}
\label{fig:big_conv_decomposition}
\end{figure*}
\section{Decomposing Convolutions}
In this section, we aim at decomposition bilinear convolution networks into an orthonormal basis of eigenfilters. 
\subsection{Tensor Decomposition of Bilinear Convolutional Tensors
}
Let us consider the output of a convolution layer at location $(\alpha,\beta)$ for the $i$-th output channel:
\begin{equation}
    u(\alpha,\beta,i)= \sum_{j}^{n} \sum_{k_1}^{K} \sum_{k_2}^{K} U^{(i)}[j, k_1, k_2] \cdot X[j, \alpha+k_1, \beta +k_2]
\end{equation}
We can define the contribution of the filter applied to the $j$th input channel as:
$$ a_j = \sum_{k_1}^{K} \sum_{k_2}^{K} U^{(i)}[j, k_1, k_2] \cdot X[j, \alpha+k_1, \beta +k_2]$$
This allows us to rewrite the output as:
\begin{equation}
    u(\alpha,\beta,i) = \sum_{j}^{n} a_j
\end{equation}
We can flatten the input tensor \( X[j, \alpha:\alpha+k, \beta:\beta+k] \) into a \(K^2 \)-dimensional vector for each spatial location \((\alpha, \beta)\). Let us denote this flattened version as \( X[j, :, :]_f \). Similarly, we can write a flattened version of the filter \( U^{(i)}[j, :, :] \), which we'll call \( U^{(i)}[j, :, :]_f \). Note that the filter is independent of the position \((\alpha, \beta)\).

Using these flattened representations, we can express $a_j$ as:
\begin{equation}
    a_j = U^{(i)}[j, :, :]_f \cdot X[j, :, :]_f
\end{equation}
For readability, we can simplify the notation of the flattened vectors. We will also remove the notation for the output channel $(i)$ , as all the operations we discuss here are for a single output channel. Simplifying the notation, we get: 
\begin{equation}
    a_j = U_j \cdot X_j
\end{equation}
Note that $U_j$ is a $K^2$ row vector, and $X_j$ is a $K^2$ column vector.
The gated operation is given by:
\[
    u(\alpha,\beta,i) \odot v(\alpha,\beta,i)
\]
where $v$ is the output of another Conv2D block.
We perform a pointwise multiplication of the outputs $u$ and $v$:
\[
u(\alpha,\beta,i) = \sum_{}^{n} a_j, \text{ where } a_j= U_j   X_j
\]
and using the same simplified notation for $V^{(i)}[k,:,:]_f$ :
\[
v[\alpha,\beta,i] = \sum_{}^{n} b_k, \text{ where } b_k = V_k X_k
\]
Therefore, for any $\alpha$ and $\beta$:
\begin{align*}
u[\alpha,\beta,i] \odot v[\alpha,\beta,i] &= \left(\sum_{j}^{n} a_j\right) \left(\sum_{k}^{n} b_k\right) \\
&= \sum_{j}^{n} \sum_{k}^{n} (a_j b_k)
\end{align*}
Interaction of channel $k$ with channel $j$ is given by:
\begin{align*}
a_j b_k
&= (U_j X_j)(V_k X_k) \\
&= ({X_j}^T{U^{T}_j})(V_k X_k) \quad \text{(since $a_j$ is a scalar)}
\end{align*}
Note that we can write the following sum:
\[
\sum_{j}^n\sum_{k}^n X_j^T U_j^{T} V_k X_k
\]
as a product of three block matrices \autoref{fig:big_conv_decomposition}:
$\begin{bmatrix} 
    X_{1}^{T} & X_{2}^{T} & \cdots & X_{n}^{T}
\end{bmatrix}
\begin{bmatrix}
U^{T}_1 V_1 & \cdots & U^{T}_1 V_n \\
\vdots & & \vdots \\
U^{T}_n V_1 & & U^{T}_n V_n
\end{bmatrix}
\begin{bmatrix} 
    X_{1} \\
    X_{2} \\
    \vdots \\
     X_{n}
\end{bmatrix}$

The bilinear matrix B has a symmetric form given by  
\[
B^{sym} = \begin{bmatrix}
\frac{U_1^{ T} V^{ }_1 + (U_1^{ T} V^{ }_1)^T}{2} & \cdots & \frac{U_1^{ T} V_n + (U_n^{ T} V_1)^T}{2} \\
\vdots & & \vdots \\
\frac{U^{ T}_n V^{ }_1 + (U_1^{ T} V_n)^T}{2} & \cdots & \frac{U^{ T}_n V_n + (U^{ T}_n V_n)^T}{2}
\end{bmatrix}
\]

We show that $B_{sym}$ is indeed symmetric over all possible input pairs:
\begin{align*}
X_j^T B^{sym} X_k &= X_j^T \left(\frac{U^{ T}_j V^{ }_k + (U_k^{ T} V^{ }_j)^T}{2}\right) X_k \\
X_k^T B^{sym} X_j &= X_k^T \left(\frac{U_k^{ T} V_j + (U_j^{ T} V_k^{ })^T}{2}\right) X_j \\
&= X_j^T \left(\frac{V^{ T}_j U^{ }_k + U_j^{ T} V_k^{ }}{2}\right) X_k \\
&= X_j^T \left(\frac{U_j^{ T} V_k^{ } + (U_k^{ T} V^{ }_j)^T}{2}\right) X_k
\end{align*}
Additionally, we can see  that:
\begin{align}
X_j^T \left(\frac{\textcolor{red}{U_j^T V_k} + \textcolor{blue}{(U_k^T V_j)^T}}{2}\right) X_k \\
+ X_k^T \left(\frac{\textcolor{blue}{U_k^T V_j} + \textcolor{red}{(U_j^T V_k)^T}}{2}\right) X_j \\
= X_j^T \textcolor{red}{U^T_j V_k} X_k + X_k^T \textcolor{blue}{U^T_k V_j} X_j 
\end{align}
for all j and k. The respective red and blue terms are compatible, because each term in the expansion is a scalar and is thus equal to its transpose. Therefore, $B^{\text{sym}}$ agrees with B on every input.

For each of $m$ output channels, we get a matrix $B^{\text{sym}}$ of dimension $nK^2 \times nK^2$, making its total shape $[nK^2,nK^2,m]$.
\subsection{Bilinear component decomposition protocol}
Similar to the decomposition approach in \citep{pearce2024bilinear}, we can fix an output vector $u \in \mathbb{R}^m$ and multiply it by $B^{sym}$ along the output channel dimension to produce a matrix $Q^u=uB^\text{sym}$ of shape $[nK^2, nK^2]$, that functions as a quadratic form on the input space. Note an important distinction: since Conv layers are not fully connected but are rather \emph{locally connected}, the output vector $u$ is in $[C_\text{out}]$ space, and the decomposition produces an eigenbasis for the filters that we call \emph{eigenfilters}. That is, you get a basis consisting of $nK^2$ \emph{eigenfilters} of shape $(K\cdot K \cdot n)$. In spectral theorem terminology, we have $Q^u=F^T\Lambda F$, where $F$ is an orthonormal matrix (satisfying $F^{-1} = F^T$) of eigenvectors, and $\Lambda$ is a real, diagonal matrix of eigenvalues. 

\subsubsection{Contributions of Eigenfilters}
Since $Q_u$ is used in practice as a quadratic form, its contributions towards $u$ for a patch $x$ centered around a given position are given by
$Q_u(x)=x^TQ_x=x^TF^T_u\Lambda_u F_ux=(F_ux)^T\Lambda_u (F_ux) = \sum_{i} \lambda_u^i(f^i_ux)^2$ and $Q_u=\sum_i\lambda_u^if^i_uf^{iT}_u$.
Each $f^i$ is an individual eigenfilter, and has shape $[(K_w \cdot K_h \cdot C_\text{in})]$. As the eigenfilter activations are applied to every valid position uniformly, we can equivalently write $Q_u(X)=\sum_{i} \lambda_u^i(f_u^i \ast X)^2$.
\subsubsection{Separating channels from spatial coordinates with SVD}
Suppose we have a weight or activation vector $X$ with shape $[w,h,C]$, having both spatial and channel dimensions. We can reshape $X$ into a matrix of shape $[C,wh]$, and apply singular value decomposition (SVD). This gives us
$$X = S\Sigma V^T=\sum_i \sigma_i s_iv_i^T$$ where $S$ has shape $[C,C]$, and $V$ has shape $[wh, wh]$. The top left singular vectors $s_i$ live in the channel space, and can be used as output vectors for a BConv layer.

Since the top singular vectors in channel space also have a singular value, we can aggregate the contributions of the eigenvalues and the eigenvectors together.
We can derive an eigendecomposition of the BConv layer for each singular channel, to get the following:
\begin{align*}
    Q^\text{probe}(X) &= \sum_{j=1}^m s_jQ_{u_j}(X) \\
    &= \sum_{j=1}^m s_j\sum_{i} \lambda_{u_j}^i(f^i_{u_j} \ast X)^2 \\
    &= \sum_{j=1}^m \sum_i (s_j \lambda^i_{u_j})(f^i_{u_j} \ast X)^2
\end{align*}
That is, the joint term $s_j \lambda^i_{u_j}$ parametrizes the importance of an eigenfilter for its singular channel. Note that these importance terms are signed, as the eigenvalues can be negative.

\begin{figure}[ht]
\centering
\includegraphics[width=0.46\textwidth]{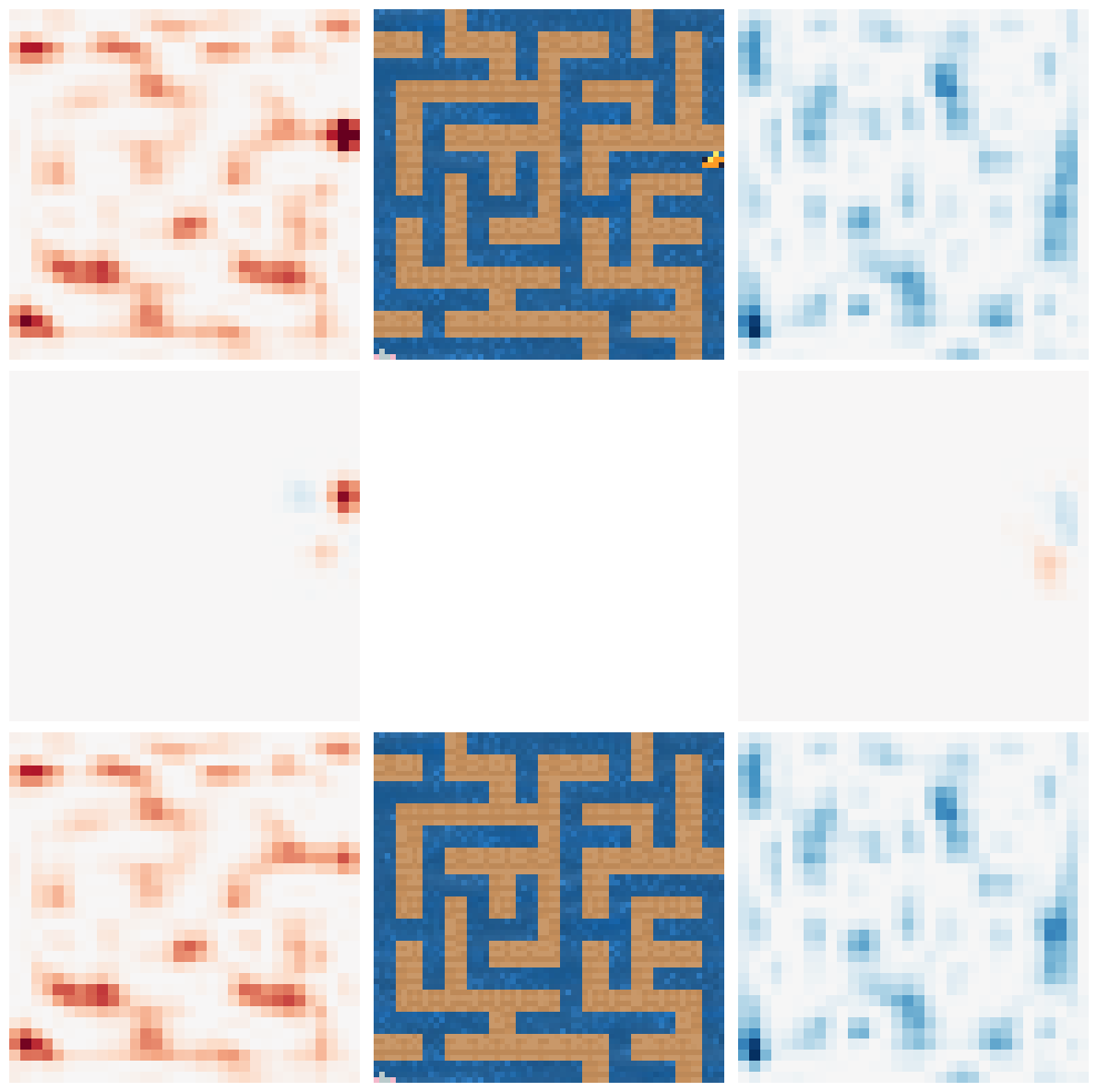}
\caption{Activations for the top positive (left) and negative (right) eigenfilters in the second BConv layer, for the cheese probe's top singular channel. Activations for a maze with cheese (top) vs without cheese (bottom). Middle plots show the difference between the activations with and without cheese. While the positive filter activates on non-cheese patterns, the negative filter downweighs non-cheese patterns without erasing the cheese activation.}
\label{fig:posneg_cheese_acts}
\end{figure}
\section{Experiments}
In this section, we evaluate the performance and interpretability of bilinear models using a series of reinforcement learning experiments and analyses. Our goal is twofold: (1) to assess whether bilinear architectures achieve competitive or superior performance compared to standard models like ReLU-based IMPALA, and (2) to explore how bilinear layers provide interpretable representations through eigendecomposition and probe-based analyses. We detail training procedures, experimental protocols, and key findings from both quantitative and qualitative perspectives.
\subsection{Architecture baseline}
The architecture of our model was adapted from an existing framework described by \cite{espeholt2018impala}. This adaptation involved modifying the original structure to incorporate bilinear gating mechanisms in both convolutional and fully connected layers. Additionally we removed some convolutional layers so that the residual block is a simple gated convolution with a skip connection.

More importantly, we replaced Conv-Relus with Bilinear Convs  and Relu FCs with Bilinear FCs, and omitting the down projection $W_2$ in $FC_{\textbf{bilinear}}$ (\autoref{fig:bimpala_architecture}).
\subsubsection{Bimpala matches performance on ProcGen tasks}

\begin{figure}[ht]
    \centering
    \includegraphics[width=0.46\textwidth]{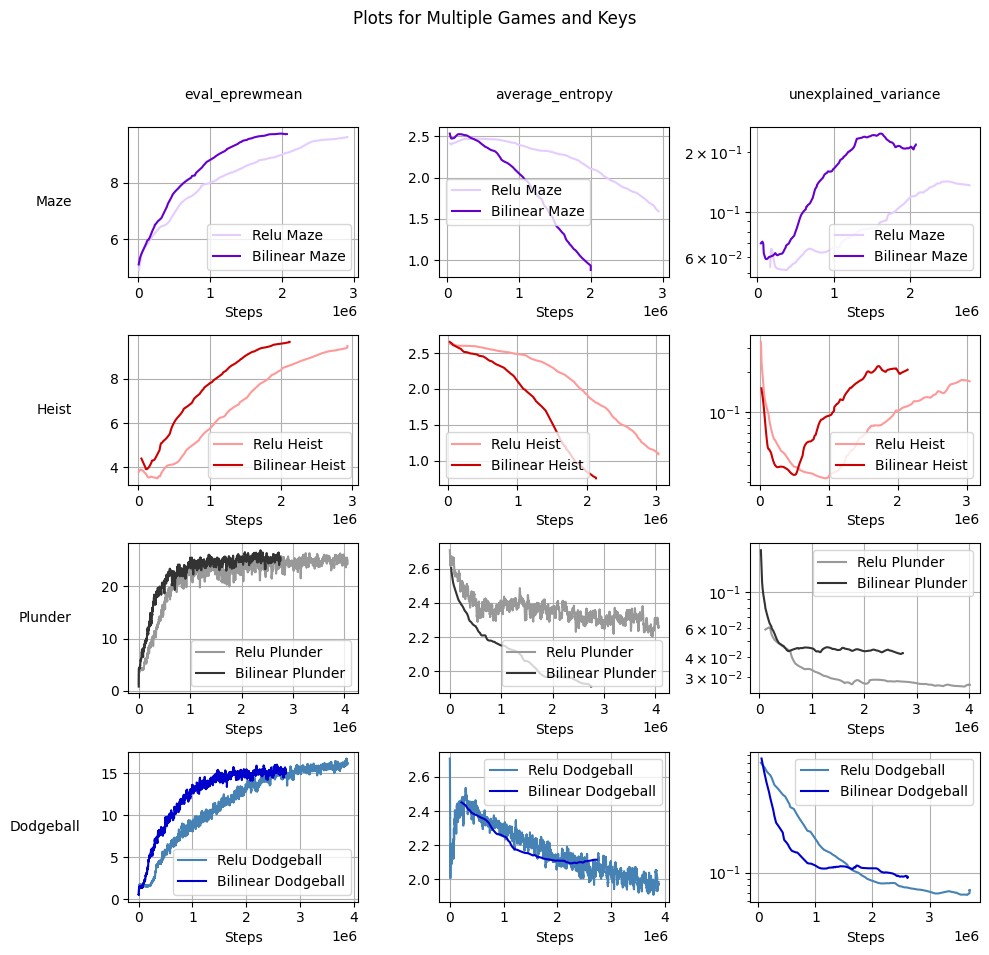}
    \caption{Performance comparison between ReLU and Bilinear architectures across four reinforcement learning environments. Each row represents a different environment (Maze, Heist, Plunder, Dodgeball), while columns show different evaluation metrics (expected return, average entropy, and unexplained variance). The Bilinear architecture (shown in darker colors) generally demonstrates faster learning and higher final performance in terms of expected return, and maintaining lower entropy. Unexplained variance in bilinear models is higher compared to the ReLU baseline (shown in lighter color).}
    \label{fig:metrics}
\end{figure}

We used the ProcGen environment \citep{cobbe2020procgen} to train reinforcement learning policy models with PPO. Specifically, we trained a simplified ReLU network alongside a bilinear variant, which we refer to as Bimpala (Bilinear IMPALA). Our results show that Bimpala matches and occasionally outperforms IMPALA across several tasks in ProcGen, including \textbf{Maze, Heist, Plunder}, and \textbf{DodgeBall}. We trained on the "easy" distribution for all these environments due to less time steps required for convergence.(\autoref{fig:metrics}).

These results validate the feasibility of using bilinear layers for reinforcement learning tasks. We now turn to describing how this architecture can be used to enhance interpretability for reinforcement learning.
\subsection{Analysis protocol}
We suggest a protocol to connect bottom-up mechanistic approaches to top-down concept based approaches.
\begin{enumerate}
    \item Train a linear probe for a concept of interest on a Conv activation space with shape $[{\rm width}, {\rm height}, C]$. Reshape as $[C, {\rm width} \cdot {\rm height}]$
    \item Rewrite the probe's weights using the SVD, and use the top left channel-space singular vectors as output directions for a preceding BConv layer. Determine the number $m$ of singular components needed, based on the distribution of singular values.
    \item Perform an eigendecomposition towards the top $m$ left singular vectors in channel space, to identify directions in the filter weights that write to the probe.
\end{enumerate}

Since we're getting a full basis of eigenvectors for each output direction, it's possible for the important eigenvectors between output directions to not be fully orthogonal. This is especially relevant if interpreting multiple probes in parallel. Analyzing the cosine similarity between important eigenvectors relating to different singular channels, where importance is measured by $|s_j\lambda^i_{u_j}|$, may further inform the function of the eigenvectors, although we do not investigate overlapping filters in this draft.

With the protocol defined, the next step is to implement it by training concept probes for specific features and analyzing their decomposition.
\subsection{Methodology}
\subsubsection{Training concept probes}
In this analysis, we use a Bimpala model trained on the ProcGen Maze environment, where the player, a mouse, must navigate a maze to find the sole piece of cheese and earn a reward. We trained linear probes to detect the presence of the cheese at position $(8,14)$ in the maze by creating a dataset comprising 2000 mazes with a cheese at position $(8,14)$ and 2000 mazes without cheese.
We see that probes trained on the outputs of the residual blocks get about $99 \%$  accuracies and $F_1$ scores (\autoref{tab:f1_scores}). 
\begin{table}[ht]
\centering
\begin{tabular}{|l|c|}
\hline
\textbf{Layer} & \textbf{F1 Score (\%)} \\
\hline
Initial Conv & 99.88 \\
\hline
\multicolumn{2}{|l|}{\textbf{Sequence 0}} \\
\hline
Conv & 100.00 \\
MaxPool & 99.88 \\
ResBlock0 & 99.88 \\
ResBlock0 Gated Conv & 59.28 \\
ResBlock1 & 100.00 \\
ResBlock1 Gated Conv & 69.07 \\

\hline
\multicolumn{2}{|l|}{\textbf{Sequence 1}} \\
\hline
Conv & 100.00 \\
MaxPool & 100.00 \\
ResBlock0 & 100.00 \\
ResBlock0 Gated Conv & 42.75 \\
ResBlock1 & 100.00 \\
ResBlock1 Gated Conv & 0.00 \\

\hline
\multicolumn{2}{|l|}{\textbf{Sequence 2}} \\
\hline
Conv & 100.00 \\
MaxPool & 100.00 \\
ResBlock0 & 100.00 \\
ResBlock0 Gated Conv & 69.07 \\
ResBlock1 & 100.00 \\
ResBlock1 Gated Conv & 40.12 \\

\hline
\multicolumn{2}{|l|}{\textbf{Fully Connected Layers}} \\
\hline
Gated FC & 68.36 \\
Logits FC & 81.20 \\
Value FC & 2.73 \\
\hline
\end{tabular}
\caption{$F_1$ scores for position probes trained on the output of different layers of the network. The scores indicate how well each layer preserves cheese position information.}
\label{tab:f1_scores}
\end{table}

\subsubsection{Dominant singular probe channels}
\begin{figure}[ht]
    \centering
    \begin{minipage}{0.23\textwidth}
        \centering
        \includegraphics[width=\textwidth]{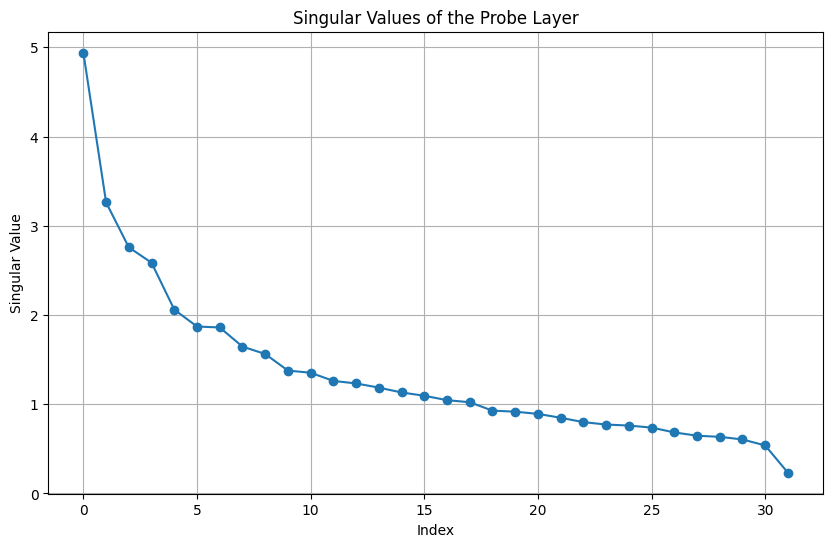}
    \end{minipage}
    \hfill
    \begin{minipage}{0.23\textwidth}
        \centering
        \includegraphics[width=\textwidth]{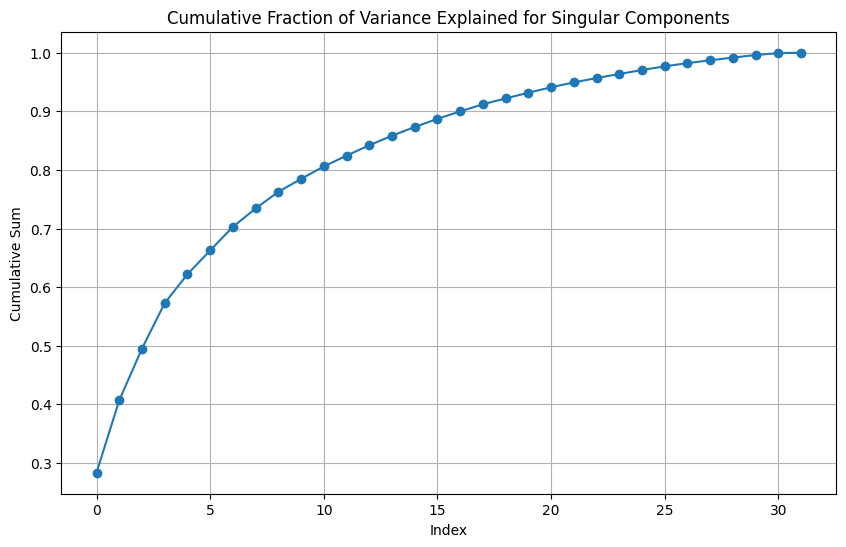}
    \end{minipage}
    \caption{Left: Probe singular values. Right: Fraction of variance explained.} 
    \label{fig:probe_singular_values} 
\end{figure}
Next, we apply singular value decomposition to the probes. The top singular component alone explains $30\%$ of the variance, and $16$ components are needed to explain $\geq 90\%$ of the variance  (\autoref{fig:probe_singular_values}).
\subsubsection{Eigenfilter decomposition for singular probe channels}
\begin{figure}[ht]
    \centering
    \begin{minipage}{0.23\textwidth}
        \centering
        \includegraphics[width=\textwidth]{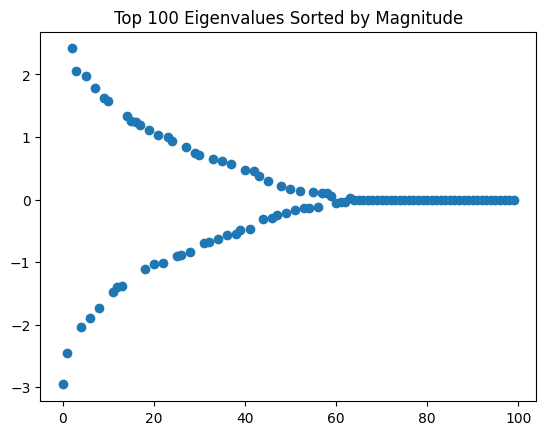}
    \end{minipage}
    \hfill
    \begin{minipage}{0.23\textwidth}
        \centering
        \includegraphics[width=\textwidth]{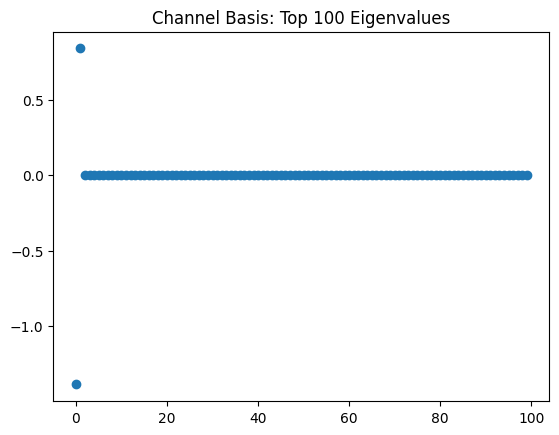}
    \end{minipage}
    
    \caption{Left: Eigenfilter spectrum towards top singular channel. Right: Eigenfilter spectrum in the standard channel basis. Singular spectrum is nondegenerate, whereas the channel basis spectra have just two eigenvalues, indicating it is not an informative basis.}
    \label{fig:eigenfilter_spectrum}
\end{figure}
We then decomposed the preceding BConv layer towards $u_1$, the top singular channel. The spectrum for singular channels is nondegenerate, whereas the basis aligned or random channel spectra have just two nonzero eigenvalues (\autoref{fig:eigenfilter_spectrum}).  
We visualize the top positive and negative eigenfilter activations for a set of pairs of mazes, one with the cheese at the selected position and the other without the cheese.
The positive and negative activations of the respective filters result in a cheese detector filter (\autoref{fig:posneg_cheese_acts}).
While the positive filter
activates on non-cheese patterns, it activates the strongest at the cheese location, and the negative filter downweighs
non-cheese patterns without erasing the cheese activation.

\subsubsection{Action features}
\begin{figure}[ht]
    \centering
    \includegraphics[width=0.46\textwidth]{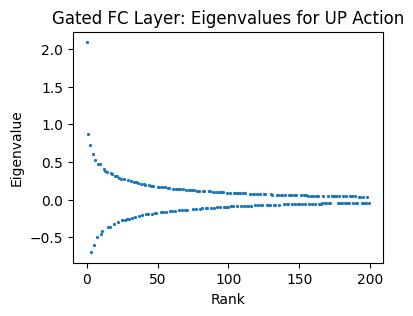} 
    \caption{Action spectrum. There are many relevant eigenvalues, with one very large positive eigenvalue.} 
    \label{fig:action_spectrum} 
\end{figure}

\begin{figure}[ht]
    \centering
    \includegraphics[width=\columnwidth]{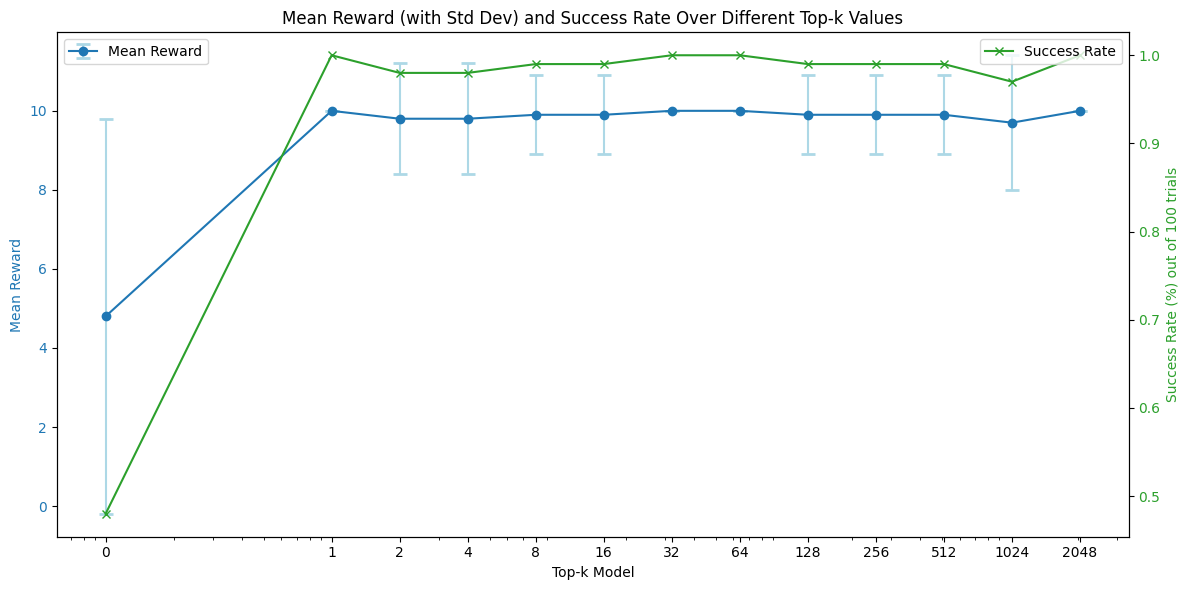} 
    \caption{Keeping just 1eigenvector for each output action is enough to preserve near 100\% success rate in solving mazes.} 
    \label{fig:topk_action_rollouts} 
\end{figure}

Instead of training probes, we could alternatively decompose the directions relevant for actions directly. There are many action eigenvectors in the final FC layer (\autoref{fig:action_spectrum}).
Interestingly, despite a dense spectrum, ablating all but the top action vector is sufficient to preserve maze-solving ability (\autoref{fig:topk_action_rollouts}).
\subsection{Ablation Studies}

\begin{figure}[ht]
    \centering
    \includegraphics[width=\columnwidth]{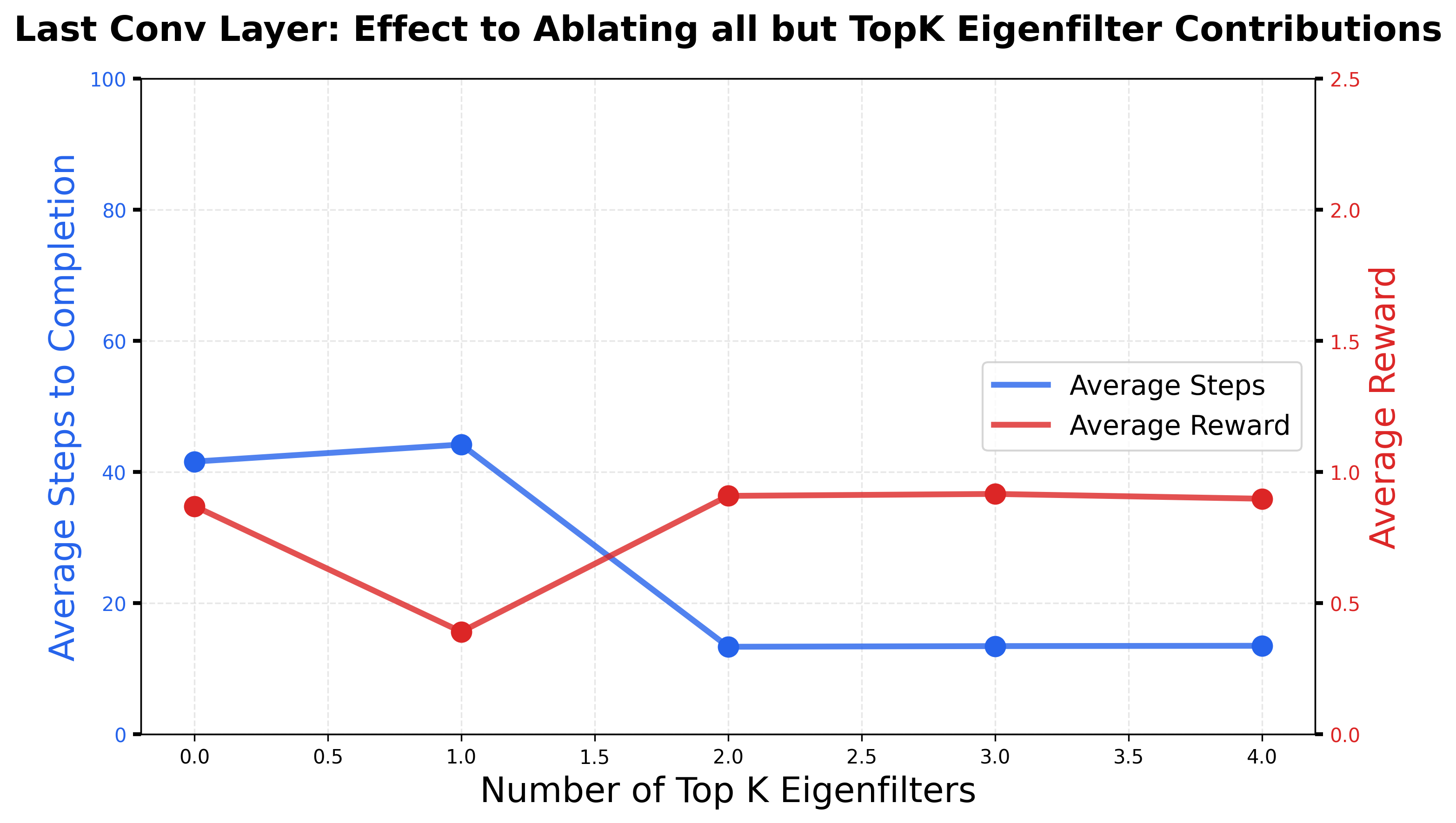} 
    \caption{Last Conv Layer: Ablating eigenvectors lead to the agent taking more steps to solve the maze. Top $2$ eigenvectors per channel is sufficient for the agent to recover its performance.} 
    \label{fig:topk_final_conv} 
\end{figure}

We carry out ablations in the standard channel basis at different layers of the network to figure out how large is the computation basis of the network. We ablate all but the top $k$ eigenfilters for each output channel of the final BConv layer. After knocking out the entire layer, maze solving ability is preserved but it takes more steps on average to solve. We cap the steps per rollout at $200$ and we use $20$ seeded environments and ensure that each set of ablations are performed on the same test set. When preserving $2$ eigenfilters per output channel, we see full performance recovered (\autoref{fig:topk_final_conv}).

\begin{figure}[ht]
    \centering
    \includegraphics[width=\columnwidth]{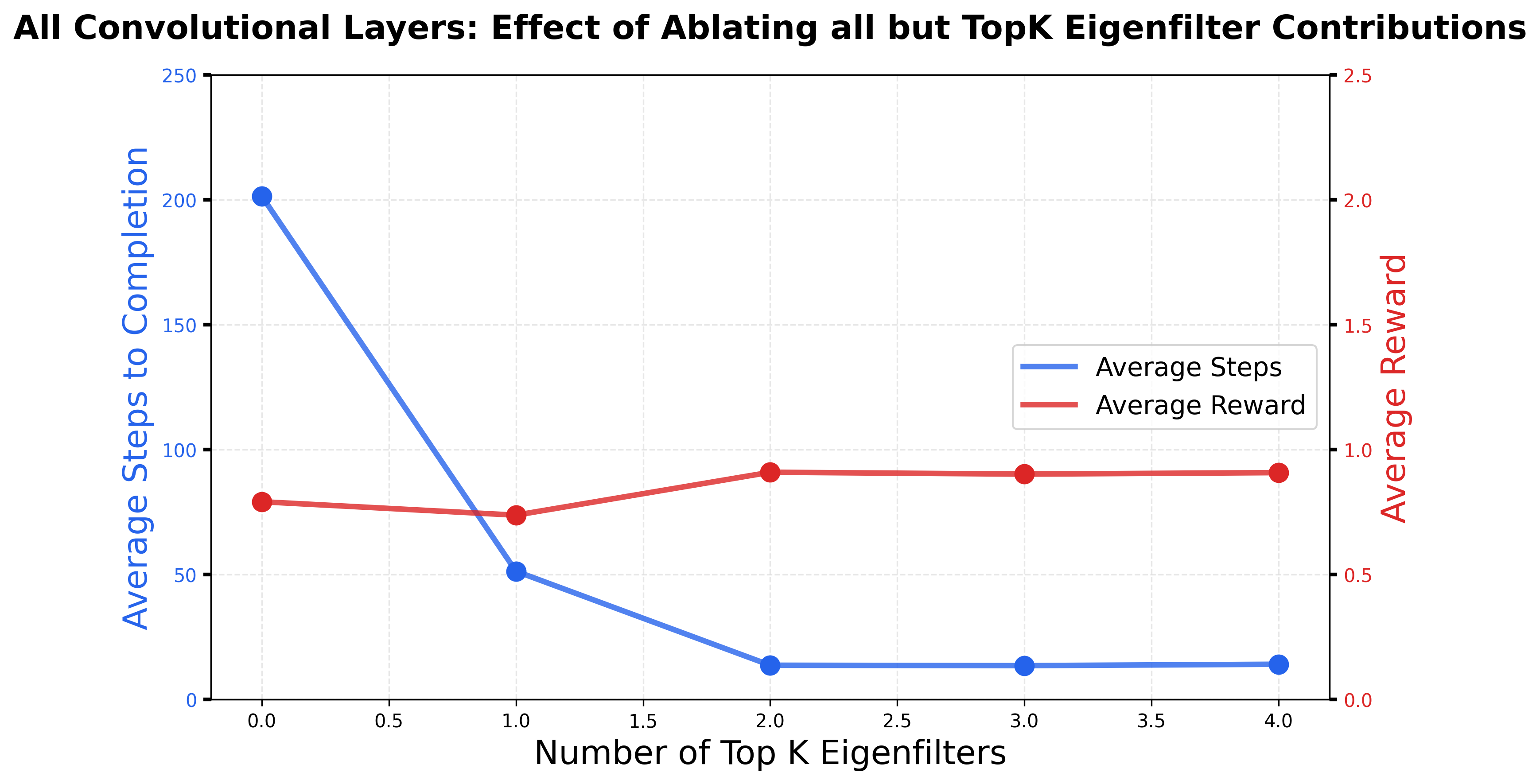} 
    \caption{Ablating all but the top $k$ eigenfilters for all output channels in the standard basis, across all layers. The top positive and negative eigenfilters must work together, but it doesn't tell us anything more granular.} 
    \label{fig:topk_all_conv} 
\end{figure}

To jointly recover performance within the maze, we compare performance of the agent where every eigenfilter other than the top $k$ eigenfilters are ablated per output channel (in the standard basis) across all layers. We find that we need $2$ eigenfilters per channel to recover performance (\autoref{fig:topk_all_conv}). With $k=1$, all convolutions are ablated, and the agent fails to solve the maze (as it reaches the max steps). With $k=1$ eigenfilters, we find that the agent does solve the maze, but it takes relatively more time compared to average as it makes some wrong moves.

\section{Discussion}
\textbf{Summary}
We introduce an approach to interpreting convolutional neural networks, by replacing nonlinearities with bilinear variants that achieve comparable and occasionally superior performance, although this was not our aim. Our approach allows us to find a closed form for self-interacting convolution features that can be combined with a top down concept based approach to derive causally relevant mechanisms used by RL agents in their decision making process. Therefore, we see great value in bilinear variants that offer more interpretability prospects while achieving competitive performance to its non-analytic variants.

\textbf{Limitations}
We found significant challenges in interpreting the units of computation in a entirely data independent fashion, and found that top activating dataset examples for eigenvectors tend to not be informative. However, the decomposition allows us to break concept probes into more granular units of computation.
We considered only one architecture, IMPALA, for our policy, although we expect the general approach of replacing nonlinearities with bilinear variants to be widely applicable.

We do not address a range of components often found in convolutional neural networks, such as batch norm, dropout, or pooling. Considering the implications for each of these, such as the performance tradeoff for linear or quadratic versus max pooling, is important for assessing the viability of architecture variants. We are looking forward to exploring the interations of eigenvectors over several layers of the network to investigate multi-step reasoning and do a lower level mechanistic interpretability study of bilinear RL variants.

\bibliography{arxiv}
\bibliographystyle{icml2024}

\end{document}